\documentclass[10pt,twocolumn,letterpaper]{article}

\usepackage{cvpr}
\usepackage[]{authblk}
\usepackage{times}
\usepackage{epsfig}
\usepackage{graphicx}
\usepackage{amsmath}
\usepackage{amssymb}
\usepackage{bm}
\usepackage{multirow}
\usepackage[tight,footnotesize]{subfigure}

\usepackage{balance}\usepackage{graphicx}
\usepackage{latexsym, amsfonts, amsmath, amsthm, amssymb, mathrsfs, bm}
\usepackage{multirow}
\usepackage{algorithm}
\usepackage{algorithmic}
\usepackage{multirow}
\usepackage[tight,footnotesize]{subfigure}

\usepackage{threeparttable}
\usepackage{amsmath}
\usepackage{dcolumn}
\usepackage{multirow}
\usepackage{booktabs}

\usepackage[breaklinks=true,bookmarks=false]{hyperref}

\cvprfinalcopy 

\def\cvprPaperID{****} 

\begin{document}

\title{Packing and Padding: Coupled Multi-index for Accurate Image Retrieval}

\author[1]{Liang Zheng}
\author[1]{Shengjin Wang}
\author[1]{Ziqiong Liu}
\author[2]{Qi Tian}
\setlength{\affilsep}{0em}\affil[1]{State Key Laboratory of Intelligent Technology and Systems;} \affil[1]{Tsinghua National Laboratory for Information Science and Technology;}\affil[1]{Department of Electronic Engineering, Tsinghua University, Beijing 100084, China}
\affil[2]{University of Texas at San Antonio, TX, 78249, USA
\authorcr{\tt\small zheng-l06@mails.tsinghua.edu.cn \tt\small wgsgj@tsinghua.edu.cn}
\authorcr{\tt\small liuziqiong@ocrserv.ee.tsinghua.edu.cn \tt\small qitian@cs.utsa.edu}}

\maketitle

\begin{abstract}
   In Bag-of-Words (BoW) based image retrieval, the SIFT visual word has a low discriminative power, so false positive matches occur prevalently. Apart from the information loss during quantization, another cause is that the SIFT feature only describes the local gradient distribution. To address this problem, this paper proposes a coupled Multi-Index (c-MI) framework to perform feature fusion at indexing level. Basically, complementary features are coupled into a multi-dimensional inverted index. Each dimension of c-MI corresponds to one kind of feature, and the retrieval process votes for images similar in both SIFT and other feature spaces. Specifically, we exploit the fusion of local color feature into c-MI. While the precision of visual match is greatly enhanced, we adopt Multiple Assignment to improve recall. The joint cooperation of SIFT and color features significantly reduces the impact of false positive matches.

   Extensive experiments on several benchmark datasets demonstrate that c-MI improves the retrieval accuracy significantly, while consuming only half of the query time compared to the baseline. Importantly, we show that c-MI is well complementary to many prior techniques. Assembling these methods, we have obtained an mAP of 85.8$\%$ and N-S score of 3.85 on Holidays and Ukbench datasets, respectively, which compare favorably with the state-of-the-arts.
\end{abstract}

\section{Introduction}

This paper considers the task of near duplicate image retrieval in large scale databases. Specifically, given a query image, our goal is to find all images sharing similar appearance in real time.

Many state-of-the-art image retrieval systems rely on the Bag-of-Words (BoW) representation. In this model, local features such as the SIFT descriptor \cite{SIFT2} are extracted and quantized to \emph{visual words} using a pre-trained \emph{codebook}. Typically, each visual word is weighted using the \emph{tf-idf} scheme \cite{video_google, zheng2013lp}. Then, an \emph{inverted index} is leveraged to reduce computational burden and memory requirements, enabling fast online retrieval.


\begin{figure}[t]
\centering
\includegraphics[width=3.18in]{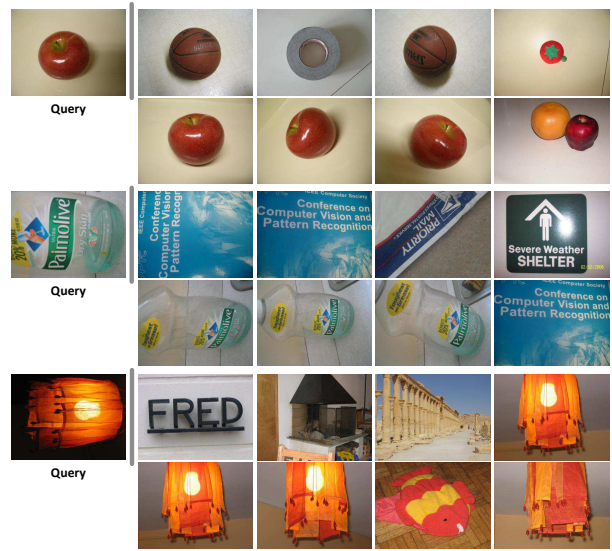}
\caption{Three examples of image retrieval from Ukbench (\textbf{Top} and \textbf{Middle}) and Holidays (\textbf{Bottom}) datasets. For each query (left), results obtained by the baseline (the first row) and c-MI (the second row) are demonstrated. The retrieval results start from the second image in the rank list.}
\label{fig:search_example}
\end{figure}

One crucial aspect in the BoW model concerns visual matching between images based on visual words. However, the reliance on the SIFT feature leads to an ignorance of other characteristics, such as color, of an image. This problem, together with the information loss during quantization, leads to many false positive matches and thus compromises the retrieval accuracy.

To enhance the discriminative power of SIFT visual words, we present a coupled Multi-Index (c-MI) framework to perform local feature fusion at indexing level. To the best of our knowledge, it is the first time that a multi-dimensional inverted index is employed in the field of image retrieval. Particularly, this paper ``couples'' SIFT and color features into a multi-index \cite{babenko2012inverted}, so that efficient yet effective image retrieval can be achieved. The final system of this paper consists of ``packing'' and ``padding'' modules.


In the ``packing'' step, we construct the coupled Multi-Index by taking each of the SIFT and color features as one dimension of the multi-index. Therefore, the multi-index becomes a joint cooperation of two heterogeneous features. Since each SIFT descriptor is coupled with a color feature, its discriminative power is greatly enhanced. On the other hand, to improve recall, Multiple Assignment (MA) is employed. Particularly, to make c-MI more robust to illumination changes, we adopt a large MA value on the side of color feature.
Fig. \ref{fig:search_example} presents three sample retrieval results of our method. We observe that c-MI improves the retrieval accuracy and returns some challenging results.

Moreover, in the ``padding'' step, we further incorporate some prior techniques to enhance retrieval performance. We show in the experiments that c-MI is well compatible with methods such as rootSIFT \cite{root_sift}, Hamming Embedding \cite{hamming}, burstiness weighting \cite{burstiness}, graph fusion \cite{zhang2012query}, etc. As another major contribution, we have achieved new state-of-the-art results on Holidays \cite{hamming} and Ukbench \cite{HKM} datasets. Namely, we obtained an mAP of 85.8$\%$ and N-S score of 3.85 on Holidays and Ukbench, respectively.

The remainder of this paper is organized as follows. After an overview of related work in Section \ref{sectioin: Related Work}, we describe the ``packing'' of c-MI framework in Section \ref{sectioin: Proposed Approach}. In Section \ref{sectioin: Experiments}, the ``padding'' methods and results are presented and discussed. Finally, we conclude in Section \ref{sectioin: Conclusion}.

\section{Related Work}
\label{sectioin: Related Work}
In the image retrieval community, a myriad of works have been proposed to improve the accuracy of image retrieval. In this section, we provide a brief review of several closely related aspects.


\emph{Matching Refinement} In visual matching, a large codebook \cite{AKM} typically means a high precision but low recall, while constructing a small codebook (e.g., 20K) \cite{jegou2010improving} guarantees high recall. To improve precision given high recall, some works explore contextual cues of visual words, such as spatial information \cite{AKM, shen2012object, zhou2013sift, contextual_weighting, chum2010unsupervised, zheng2013visual}. To name a few, Shen et al. \cite{shen2012object} perform image retrieval and localization simultaneously by a voting-based method. Alternatively, Wang et al. \cite{contextual_weighting} weight visual matching based on the local spatial context similarity. Meanwhile, the precision of visual matching can be also improved by embedding binary features \cite{hamming, BOC, scalar, liu2014cross}. Specifically, methods such as Hamming Embedding \cite{hamming} rebuild the discriminative ability of visual words by projecting SIFT descriptor into binary features. Then, efficient \emph{xor} operation between binary signatures is employed, providing further check of visual matching.

\emph{Feature Fusion} The fusion of multiple cues has been proven to be effective in many tasks \cite{shahbaz2012color, topic_modeling, niu2012context}. Since the SIFT descriptor used in most image retrieval systems only describes the local gradient distribution, feature fusion can be performed to capture complementary information. For example, Wengert et al. \cite{BOC} embed local color feature into the inverted index to provide local color information. To perform feature fusion between global and local features, Zhang et al. \cite{zhang2012query} combine BoW and global features by graph fusion and maximizing weighted density, while co-indexing \cite{co_indexing} expands the inverted index according to global attribute consistency.

\emph{Indexing Strategy} The inverted index \cite{video_google} significantly promotes the efficiency of BoW based image retrieval. Motivated from text retrieval framework, each entry in the inverted index stores information associated with each indexed feature, such as image IDs \cite{AKM, co_indexing}, binary features \cite{hamming, BOC}, etc. Recent state-of-the-art works include joint inverted index \cite{joint_index} which jointly optimizes all visual words in all codebooks. The closest inspiring work to ours includes the inverted multi-index \cite{babenko2012inverted} which addresses NN search problem by ``de-composing'' the SIFT vector into different dimensions of the multi-index. Our work departs from \cite{babenko2012inverted} in two aspects. First, the problem considered in this paper consists in the indexing level feature fusion, applied in the task of large scale image retrieval. Second, we actually ``couple'' different features into a multi-index, after which the ``coupled Multi-Index (c-MI)'' is named.
\section{Proposed Approach}
\label{sectioin: Proposed Approach}
This section gives a formal description of the proposed c-MI framework.

\subsection{Conventional Inverted Index Revisit}
\label{sectioin: 1-D inverted file}
\begin{figure}[t]
  \centering
  \includegraphics[width=3.25in]{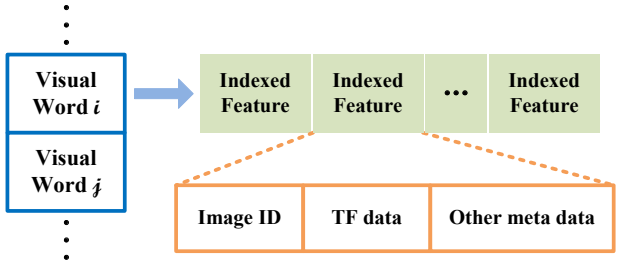}
  \caption{Conventional 1-D inverted Index. Only one kind of feature (typically the SIFT feature) is used to build the inverted Index. }
  \label{fig:1D inverted file}
\end{figure}

A majority of works in the BoW based image retrieval community employ a ONE-dimensional inverted Index \cite{zheng2013lp, AKM, HKM}, in which each entry corresponds to a visual word defined in the codebook of SIFT descriptor. Assume that a total of $N$ images are contained in an image database, denoted as $\mathcal{D} = \{I_i\}_{i=1}^N$. Each image $I_i$ has a set of local features $\{x_j\}_{j=1}^{d_i}$, where $d_i$ is the number of local features. Given a codebook $\{w_i\}_{i=1}^K$ of size $K$, a conventional 1-D inverted index is represented as $\mathcal{W} = \{W_1, W_2, ..., W_K\}$. In $\mathcal{W}$, each entry $W_i$ contains a list of indexed features, in which image ID, TF score, or other metadata \cite{hamming, zhou2013sift, contextual_weighting} are stored. An example of the conventional inverted Index is illustrated in Fig. \ref{fig:1D inverted file}.

Given a query feature, an entry $W_i$ in the inverted index is identified after feature quantization. Then, the indexed features are taken as the candidate nearest neighbors of the query feature.
In this scenario, the matching function $f_q(\cdot)$ of two local features $x$ and $y$ is defined as
\begin{equation}\label{eq: matching function 1D}
  f_q(x, y) = \delta_{q(x), q(y)},
\end{equation}
where $q(\cdot)$ is the quantization function maping a local feature to its nearest centroid in the codebook, and $\delta$ is the Kronecker delta response.

The 1-D inverted index votes for candidate images similar to the query in \emph{one} feature space, typically the SIFT descriptor space. However, the intensity-based features are unable to capture other characteristics of a local region. Moreover, due to the quantization artifacts, the SIFT visual word is prone to producing false positive matches: local patches, similar or not, may be mapped to the same visual word. Therefore, it is undesirable to take visual word as the only ticket to feature matching. While many previous works use spatial contexts \cite{zhou2013sift, contextual_weighting} or binary features \cite{hamming} to filter out false matches, our work, instead, proposes to incorporate local color feature to provide additional discriminative power via the coupled Multi-Index (c-MI).

\subsection{Feature Extraction and Quantization}
\label{section: quantization}
\begin{figure}[t]
\centering
\includegraphics[width=3.25in]{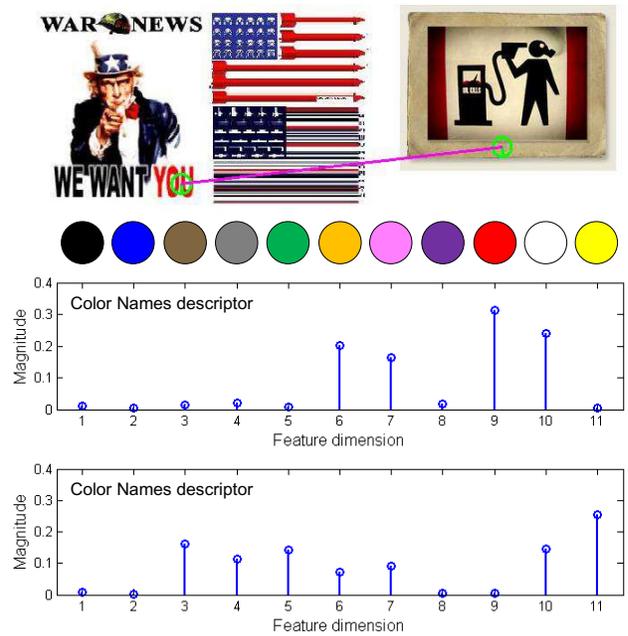}
\caption{An example of visual match. \textbf{Top:} A matched SIFT pair between two images. The Hamming distance between their 64-D SIFT Hamming signatures is 12. The 11-D color name descriptors of the two keypoints in the left (\textbf{middle}) and right (\textbf{bottom}) images are presented below. Also shown are the prototypes of the 11 basic colors (colored discs). In this example, the two local features are considered as a good match both by visual word equality and Hamming distance consistency. However, they differ a lot in color space, thus considered as a false positive match in c-MI. }
\label{fig:visual_match}
\end{figure}

This paper considers the coupling of SIFT and color features. The primary reason lies in that feature fusion works better for features with low correlation, such as SIFT and color. In feature matching, complementary information may be of vital importance. For example, given two keypoints quantized to the same SIFT visual word, if the coupled color features are largely different, they may be considered to be a false match (see Fig. \ref{fig:visual_match} for an illustration). To this end, SIFT and color features are extracted and subsequently quantized as follows.

\textbf{SIFT extraction}: Scale-invariant keypoints are detected with detectors, e.g. DoG \cite{SIFT2}, Hessian-affine \cite{AKM}, etc. Then, a 16$\times$16 patch around each keypoint is considered, from which a 128-dimensional SIFT vector is calculated.

\textbf{Color extraction}: we employ the Color Names (CN) descriptor \cite{shahbaz2012color}. CN assigns a 11-D vector to each pixel, in which each entry encodes one of the eleven basic colors: black, blue, brown, grey, green, orange, pink, purple, red, white, and yellow. Around each detected keypoint, we consider a local patch with an area proportional to the scale of the keypoint. Then, CN vectors of each pixel in this area are calculated. We take the mean CN vector as the color descriptor coupling SIFT for the current keypoint.

\textbf{Quantization} For SIFT and CN descriptors, we use the conventional quantization scheme as in \cite{AKM}. Codebooks are trained using independent SIFT and CN descriptors, respectively. Each descriptor is quantized to the nearest centroid in the corresponding codebook by Approximate Nearest Neighbor (ANN) algorithm. To improve recall, Multiple Assignment (MA) is applied. Particularly, to deal with the illumination variations, MA is set large for CN feature.

\textbf{Binary signature calculation} In order to reduce quantization error, we calculate binary signatures from original descriptors. For a SIFT descriptor, we follow the method proposed in \cite{hamming}, resulting in a 64-D binary signature.

Nevertheless, on the side of color feature, since each dimension of the CN descriptor has explicit semantic meaning, we employ the binarization scheme introduced in \cite{scalar}. Specifically, given a CN descriptor represented as $(f_1, f_2, ..., f_{11})^\mathrm{ T }$, a 22-bit binary feature $\bm b$ can be produced as follows
\begin{equation}
(b_i, b_{i+11})=\begin{cases}
(1, 1), &\mbox{if $f_i > \hat{th_1}$}, \\
(1, 0), &\mbox{if $\hat{th_2} < f_i \leq \hat{th_1}$}, \\
(0, 0), &\mbox{if $f_i \leq \hat{th_2}$}
\end{cases}
\label{eq:quant_22bit}
\end{equation}
where $b_i(i = 1, 2,..., 11)$ is the $i$th entry of the resulting binary feature $\bm b$. Thresholds $\hat{th_1} = g_{2}$, $\hat{th_2} = g_{5}$, where $(g_1, g_2,...,g_{11})^\mathrm{ T }$ is the sorted vector of $(f_1, f_2,...,f_{11})^\mathrm{ T }$ in descending order.

\subsection{Coupled Multi-Index}
\label{section: 2-D inverted file}
\textbf{Structure of c-MI}
In \cite{babenko2012inverted}, the 128-D SIFT descriptor is \emph{de-composed} into several blocks produced by product quantization \cite{jegou2011product}. The multi-index is thus organized around the codebooks of corresponding blocks. Their approach enables more accurate nearest neighbor (NN) search for SIFT features. In our work, however, we consider the task of image retrieval, which differs from pure NN search. Moreover, contrary to \cite{babenko2012inverted} we \emph{couple} different features into a multi-index, so that feature fusion is performed at indexing level. In this paper, we consider the 2-D inverted index, which is also called \emph{second-order} in \cite{babenko2012inverted}.

\begin{figure}[t]
  \centering
  \includegraphics[width=3.25in]{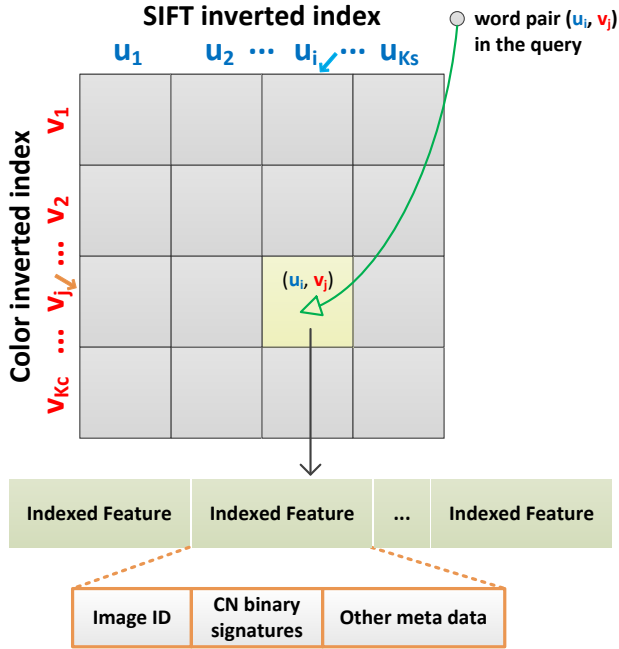}
  \caption{Structure of c-MI. The codebook sizes are $K_s$ and $K_c$ for SIFT and color features, respectively. During online retrieval, the entry of word tuple $(u_i, v_j)$ is checked.}
  \label{fig:multi_index}
\end{figure}

Let $\vec x = [x^s, x^c] \in \mathcal{R}^{D_{s+c}}$ be a coupled feature descriptor at keypoint $p$, where $x^s \in \mathcal{R}^{D_s}, x^c \in \mathcal{R}^{D_c}$ are SIFT and color descriptors of dimension $D_s$ and $D_c$, respectively. For c-MI, two codebooks are trained for each feature. Specifically, for SIFT and color descriptors, codebooks $\mathcal{U} = \{u_1, u_2, ..., u_{K_s}\}$ and $\mathcal{V} = \{v_1, v_2, ...,v_{K_c}\}$ are generated, where $K_s$ and $K_c$ are codebook sizes, respectively. As a consequence, c-MI consists of $K_s \times K_c$ entries, denoted as $\mathcal{W} = \{W_{11}, W_{12}, ..., W_{ij}, ...,  W_{K_s K_c}\}$, $i = 1, ..., K_s$, $j = 1, ..., K_c$, as illustrated in Fig. \ref{fig:multi_index}.

When building the multi-index, all feature tuples $\vec x = [x^s, x^c]$ are quantized into visual word pairs $(u_i, v_j), i = 1, ..., K_s, j = 1, ..., K_c$ using codebooks $\mathcal{U}$ and $\mathcal{V}$, so that $u_i$ and $v_j$ are the nearest centroids to features $x^s$ and $x^c$ in codebooks $\mathcal{U}$ and $\mathcal{V}$, respectively. Then, in the entry $W_{ij}$, information (e.g. image ID, CN binary signatures and other meta data) associated with the current feature tuple $\vec x$ is stored continuously in memory.

\textbf{Querying c-MI }
Given a query feature tuple $\vec x = [x^s, x^c]$, we first quantize it into a visual word pair $(u_i, v_j)$ as in the offline phase. Then, the corresponding entry $W_{ij}$ in c-MI is identified, and the list of indexed features are taken as the candidate images, similar to the classic inverted index described in Section \ref{sectioin: 1-D inverted file}. In essence, the matching function $f_{q_s, q_c}^0(\cdot)$ of two local feature tuples $\vec x = [x^s, x^c]$ and $\vec y = [y^s, y^c]$ is written as
\begin{equation}\label{eq: matching function 2D}
  f_{q_s, q_c}^0(\vec x, \vec y) = \delta_{q_s( x^s), q_s( y^s)} \cdot \delta_{q_c( x^c), q_c( y^c)},
\end{equation}
where $q_s(\cdot)$ and $q_c(\cdot)$ are quantization functions for SIFT and CN features, respectively, and $\delta$ is the Kronecker delta response as in Eq. \ref{eq: matching function 1D}. As a consequence, a local match is valid only if the two feature tuples are similar both in SIFT and color feature spaces.

Moreover, the Inverse Document Frequency (IDF) scheme can be applied in the multi-index directly. Specifically, the IDF value of entry $W_{ij}$ is defined as
\begin{equation}\label{eq: IDF_2D}
  idf{(i, j)} = \frac{N}{n_{ij}},
\end{equation}
where $N$ is the total number of images in the database, and $n_{ij}$ encodes the number of images containing the visual word pair $(u_i, v_j)$. Furthermore, the $l_2$ normalization can be also adopted in the 2-D case. Let an image be represented as a 2-D histogram $\{h_{i, j}\}$, $i = 1, ..., K_s$, $j = 1, ..., K_c$, where $h_{i, j}$ is the term-frequency (TF) of visual word pair $(u_i, v_j)$ in image $I$, the $l_2$ norm is calculated as,
\begin{equation}\label{eq: l2norm}
  \|  I  \|_2 = \left(\sum_{i=1}^{K_s} \sum_{j=1}^{K_c} h_{i, j}^2\right)^{\frac{1}{2}}.
\end{equation}
Since our multi-index structure mainly works by achieving high precision, we employ Multiple Assignment (MA) to improve recall. To address illumination variations, we set a relatively large value to the color feature. In our experiments, we find that  $l_2$-normalization produces slightly higher performance than Eq. \ref{eq: l2norm}, which is probably due to the asymmetric structure of the coupled multi-index.

Furthermore, to enhance the discriminative power of CN visual words, we incorporate color Hamming Embedding ($\mbox{HE}^{c}$) into c-MI. Two feature tuples are considered as a match \emph{iff} Eq. \ref{eq: matching function 2D} is satisfied \emph{and} the Hamming distance $d_b$ between their binary signatures is below a pre-defined threshold $\kappa$. The matching strength is defined as $\exp(-\frac{d_b^2}{\sigma^2})$. Therefore, the matching function in Eq. \ref{eq: matching function 2D} is updated as

\begin{equation}\label{eq: matching function 2D_updated}
f_{q_s, q_c}(\vec x, \vec y) =
\begin{cases}
f_{q_s, q_c}^0(\vec x, \vec y)\cdot\exp\left(-\frac{d_b^2}{\sigma^2}\right),&d_b < \kappa, \\0,
&\mbox{otherwise.}
\end{cases}
\end{equation}
Then, in the framework of c-MI, the similarity score between a database image $I$ and query image $Q$ is defined as
\begin{equation}\label{eq: sim_score}
  sim(Q, I) = \frac{\sum_{\vec x\in Q, \vec y \in  I}f_{q_s, q_c}(\vec x, \vec y)\cdot idf^2}{\|   Q  \|_2 \| I \|_2},
\end{equation}

\section{Experiments}
\label{sectioin: Experiments}
In this section, we evaluate the proposed method on five public available datasets: the Ukbench \cite{HKM}, Holidays \cite{hamming}, DupImage \cite{zhou2013sift}, Mobile \cite{contextual_weighting} and MIR Flickr 1M \cite{huiskes08}.
\subsection{Datasets}
\textbf{Ukbench} A total of 10200 images are contained in this dataset, divided into 2550 groups. Each image is taken as the query in turn. The performance is measured by the average recall of the top four ranked images, referred to as N-S score (maximum 4).

\textbf{Holidays} This dataset consists of 1491 images from personal holiday photos. There are 500 queries, most of which have 1-2 ground truth images. mAP (mean average precision) is employed to measure the retrieval accuracy.

\textbf{DupImages} This dataset is composed of 1104 images divided into 33 groups of partial-duplicate images. 108 images are selected as queries, and mAP is again used as the accuracy measurement.

\textbf{Mobile} The Mobile dataset has 400 database images and 2500 queries, captured by mobile devices. The Top-1 ($\tau_{1}$) and Top-10 ($\tau_{10}$) precision are employed.

\textbf{MIR Flickr 1M} This is a distractor dataset, with one million images randomly retrieved from Flickr. We add this dataset to test the scalability of our method.

\subsection{Experiment Settings}
\label{section: experiment_settings}
\textbf{Baseline} This paper adopts the baseline in \cite{AKM, hamming}. Hessian Affine detector and SIFT descriptor are used for feature extraction. Following \cite{root_sift}, rootSIFT is used on every point since it is shown to be effective under Euclidean distance. We also adopt the average IDF defined in \cite{zheng2013lp} in place of the original IDF. Then, a codebook of size 20K is trained using the independent Flickr60k data released in \cite{hamming}.

\textbf{SIFT Hamming Embedding} The 64-bit SIFT Hamming Embedding ($\mbox{HE}^{s}$) \cite{burstiness} is used in addition to c-MI. The Hamming threshold is set to 30, and the weighting parameter is set to 16. Moreover, we employ the SIFT Multiple Assignment ($\mbox{MA}^{s}$) \cite{hamming} scheme on the query side, in which a SIFT descriptor is assigned to 3 visual words.

\textbf{Graph Fusion} As a post-processing step, we implemented the graph fusion algorithm proposed in \cite{zhang2012query}. We extract 1000-D global HSV histogram for each image, followed by $L_1$ normalization and square scaling, similar to rootSIFT \cite{root_sift}. Rank lists obtained by c-MI and the HSV histogram are merged, yielding new ranking results.
\begin{figure}
  \centering
  \includegraphics[width=3.03in]{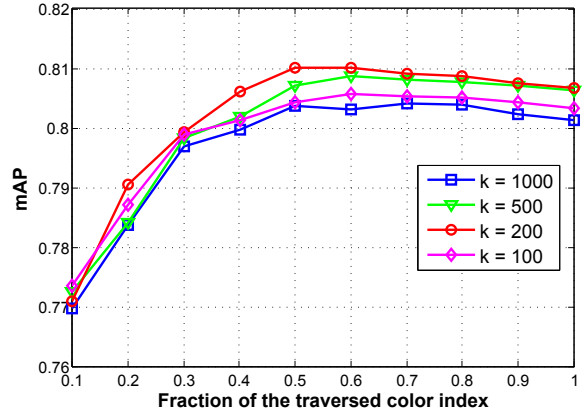}
  \caption{Impact of color codebook size on the Holidays dataset. Color codebooks of size $k = $ 100, 200, 500, and 1000 are trained on independent data. The horizontal axis represents the fraction of the codebook traversed during MA$^c$. We observe a superior performance of the codebook of size 200, and with MA$^c$ $= 200\times0.5 = 100$. Note that the query time is halved accordingly.}
  \label{fig:codebook_size}
\end{figure}

\begin{figure}
  \centering
  \includegraphics[width=3.28in]{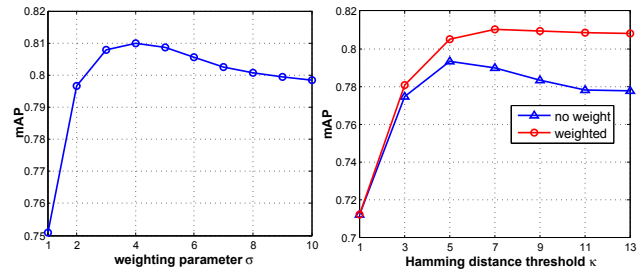}
  \caption{Influence of weighting parameter $\sigma$ (\textbf{left}) and Hamming distance threshold $\kappa$ (\textbf{right}) on Holidays dataset. mAP results are presented. We set $\sigma$ and $\kappa$ to 4 and 7, respectively.}
  \label{fig:HE_parameter}
\end{figure}

\setlength{\tabcolsep}{7.5pt}
\begin{table*}[!t]
\begin{center}
\begin{tabular}{|c|c|c|c|c||cc|c|c|c|c|}
\hline
\multirow{2}{*}{Methods} &
\multirow{2}{*}{c-MI} &
\multirow{2}{*}{$\mbox{Burst}^{s}$} &
\multirow{2}{*}{$\mbox{HE}^{s}$} &
\multirow{2}{*}{$\mbox{MA}^{s}$} &
\multicolumn{2}{c|}{\emph{Ukbench}} &
\multicolumn{1}{c|}{\emph{Holidays}} &
\multicolumn{1}{c|}{\emph{DupImage}} &
\multicolumn{2}{c|}{\emph{Mobile}}\\

\cline{6-11}
&  &  &  & & N-S & mAP(\%) & mAP(\%) &    mAP(\%) & $\tau_1$(\%)  &$\tau_{10}$(\%)\\
\hline
\hline

BoW &            &            &            &            &$3.11$     & 78.89 & $50.10$        & $48.61$    & 52.88 & 80.08\\
\hline
BoW & ${\times}$ &            &            &            &$3.43$     & 88.45 & $63.32$        & $56.28$    & 67.20 & 86.64\\
\hline
BoW & ${\times}$ & ${\times}$ &            &            &$3.52$     & 90.35 & $66.38$        & $59.87$    & 71.24 & 87.76\\
\hline
BoW & ${\times}$ &            & ${\times}$ &            &$3.64$     & 92.96 & $81.00$        & $85.10$    & 94.60 & 98.56\\
\hline
BoW & ${\times}$ & ${\times}$ & ${\times}$ &            &$3.69$     & 94.00 & $82.14$        & $86.63$    & \bf 96.48 & 98.64\\
\hline
BoW & ${\times}$ & ${\times}$ & ${\times}$ & ${\times}$ &$\bf 3.71$ & \bf 94.66 &$\bf 84.02$ & $\bf 87.60$& 96.04 & \bf98.76\\
\hline
\end{tabular}
\caption{Results for four datasets and for different methods: coupled Multi-Index (c-MI), burstiness weighting (Burst$^s$), SIFT Hamming Embedding (HE$^s$), and Multiple Assignment (MA$^s$). Parameters are selected as in Section \ref{section: experiment_settings} and \ref{section: parameter}.}
\label{table:various_approaches}
\end{center}
\end{table*}

\subsection{Parameter Analysis}
\label{section: parameter}
\textbf{Color Codebook Size and MA} We extract CN descriptors from independent images and train codebooks of various sizes, i.e., $k=100, 200, 500, 1000$. During color quantization Multiple Assignment ($\mbox{MA}^{c}$) is employed, and we vary the number of assigned words as a percentage of the codebook size. We present the mAP results on Holidays dataset in Fig. \ref{fig:codebook_size}. It is shown that the codebook of size 200 performs favorably. Moreover, assigning $50\%$ visual words in the codebook has a better performance, so  $\mbox{MA}^{c}=200\times 50\%=100$ for color feature. Intuitively, since large variation in color is often observed due to illumination, a large value of $\mbox{MA}^{c}$ helps to improve recall. Also note that ($\mbox{HE}^{s}$) with default parameters is used in parameter analysis.

\textbf{Color Hamming Embedding} Two parameters are involved in color Hamming Embedding: the Hamming distance threshold $\kappa$ and weighting factor $\sigma$. Fig. \ref{fig:HE_parameter} demonstrates the mAP results on Holidays dataset obtained by varying the two parameters. In Fig. \ref{fig:HE_parameter}(a), the mAP first rises to the peak at $\sigma = 4$ and then slowly drops. From Fig. \ref{fig:HE_parameter}(b), the best performance is achieved at $\kappa = 7$, no matter the weighted distance is employed or not. Therefore, we set $\sigma$ and $\kappa$ to 4 and 7, respectively.

\subsection{Evaluation}

\textbf{To what extent does it improve the baseline?} Implemented as described in Section \ref{section: experiment_settings}, the baseline results for Ukbench and Holidays are 3.11 in N-S score and $50.1\%$ in mAP, respectively, both higher than the reported results \cite{hamming, burstiness}. After expanding inverted index into the 2-D case as c-MI, large improvement over the baseline approach can be seen from Table \ref{table:various_approaches}. On Ukbench, we observe a big improvement of +0.32 in N-S score. Similarly on Holidays and Mobile datasets, the improvement is +13.2$\%$ in mAP and +14.3$\%$ in Top-1 precision. Note that the improvement is less prominent for DupImage (mAP from 48.6$\%$ to 56.3$\%$), because the ground truth images in this dataset have a large variety in color (even gray-level images).
\setlength{\tabcolsep}{3pt}
\begin{table}[t]
\renewcommand{\arraystretch}{1.0}
\centering
\begin{tabular}{|c|cccc|}
\hline
Methods& Ours &  HSV & HSV* & Ours + HSV* \\

\hline
\hline
\emph{Ukbench}, N-S&  3.71 & 2.97 & 3.40 &3.85\\
\hline
\emph{Holidays}, mAP(\%)&  84.02 & 59.43 & 65.29 & 85.76   \\
\hline
\end{tabular}
\caption{Performance of our method combined with graph fusion on Ukbench and Holidays datasets. * denotes results obtained by HSV histogram scaled as described in Section \ref{section: experiment_settings}.}
\label{table:graph_fusion}
\end{table}

\textbf{Complementarity to some existing methods} To test whether c-MI is compatible with some prior arts used in the 1-D inverted index, we further ``pad'' burstiness weighting ($\mbox{Burst}^{s}$) \cite{burstiness}, Hamming Embedding ($\mbox{HE}^{s}$) \cite{hamming}, Multiple Assignment ($\mbox{MA}^{s}$) \cite{hamming}, etc., into our framework. Note that these techniques are applied on the SIFT side.

It is clear from Table \ref{table:various_approaches} that these methods bring about consistent improvements. Taking Ukbench for example, the combination of $\mbox{Burst}^{s}$ and $\mbox{HE}^{s}$ each improves the N-S from 3.43 to 3.52, and from 3.43 to 3.64, respectively. Combining the three steps brings the result to 3.69. Then, the use of $\mbox{MA}^{s}$ obtains the N-S of 3.71. Similar situation is observed for the other datasets. These demonstrate the feasibility of c-MI as a general framework for image retrieval.

In addition, we add a post-processing step, i.e.,, the graph fusion of global HSV histogram \cite{zhang2012query} to the Holidays and Ukbench datasets. Similar to the scaling method applied in rootSIFT \cite{root_sift}, we also normalize the HSV histogram by its $l_1$ norm, and then exert a square root scaling. With the modified HSV histogram (HSV*), we have obtained a much higher result of HSV-based image retrieval (see Table \ref{table:graph_fusion}). After merging the graphs constructed from the c-MI and HSV rank lists, the final result arrives at 3.85 for Ukbench and 85.8$\%$ for Holidays, respectively.

\textbf{Large-scale experiments}
To test the scalability of our method, the four benchmark datasets are merged with various fractions of the MIR Flickr 1M images. In what follows, we mainly report three related aspects, i.e., accuracy, time efficiency, and memory cost.

\setlength{\tabcolsep}{11pt}
\begin{table}[t]
\renewcommand{\arraystretch}{1.0}
\centering
\begin{tabular}{|c|ccc|}
\hline
Methods& \multirow{1}{*}{Baseline}  & \multirow{1}{*}{$\mbox{HE}^{s}$} & \multirow{1}{*}{c-MI + $\mbox{HE}^{s}$} \\

\hline
\hline
\emph{Ukbench}  &   2.282  &  1.933 &   \bf 1.339\\
\hline
\emph{Holidays} &   2.722  &  2.140  &  \bf 1.413\\
\hline
\emph{DupImage} &   1.900  &  1.391  &  \bf 0.885\\
\hline
\emph{Mobile}   &   1.421  &  1.185  &  \bf 0.667\\
\hline
\end{tabular}
\caption{Average query time (s) on Ukbench, Holidays, DupImage, and Mobile + MIR Flickr 1M datasets.}
\label{table:search_time}
\end{table}

\setlength{\tabcolsep}{3.5pt}
\begin{table}[t]
\centering

\begin{tabular}{|c|cccc|}
\hline
Methods& \multirow{1}{*}{Baseline} & \multirow{1}{*}{c-MI} & \multirow{1}{*}{$\mbox{HE}^{s}$} & \multirow{1}{*}{c-MI + $\mbox{HE}^{s}$} \\

\hline
\hline
Per feature (bytes)  &  4   &  6.75  & 12 & 14.75   \\
\hline
1M dataset (GB) & 1.7      &  2.8  &  5.0  & 6.1\\
\hline

\end{tabular}
\caption{Memory cost for different approaches.}
\label{table:memory}
\end{table}

\setlength{\tabcolsep}{8pt}
\begin{table*}[t]
\centering
\begin{tabular}{|c|ccccccccccc |}
\hline
Methods& Ours & \cite{selective_match}&\cite{contextual_weighting} & \cite{jegou2010accurate} & \cite{shen2012object} & \cite{co_indexing} & \cite{BOC} & \cite{jegou2010improving} & \cite{qin2013query} & \cite{Bayes_merging} & \cite{burstiness}\\

\hline
\hline
\emph{Ukbench}, N-S score& \bf 3.71    & - &  3.56  & 3.61 &  3.52  & 3.60 & 3.50  & 3.42 & - &3.62 & 3.54\\
\hline
\emph{Holidays}, mAP(\%)& \bf 84.0    & 82.2 &  78.0  & - &  76.2  & 80.9  & 78.9 & 81.3 & 82.1 & 81.9 & 83.9\\
\hline
\end{tabular}
\caption{Performance comparison with state-of-the-art methods without post-processing}
\label{table:state_of_art_no_post}
\end{table*}

\begin{figure} [t]
\centering
\subfigure[Ukbench]{\label{fig:ukbench_scale}%
\includegraphics[width=1.6in]{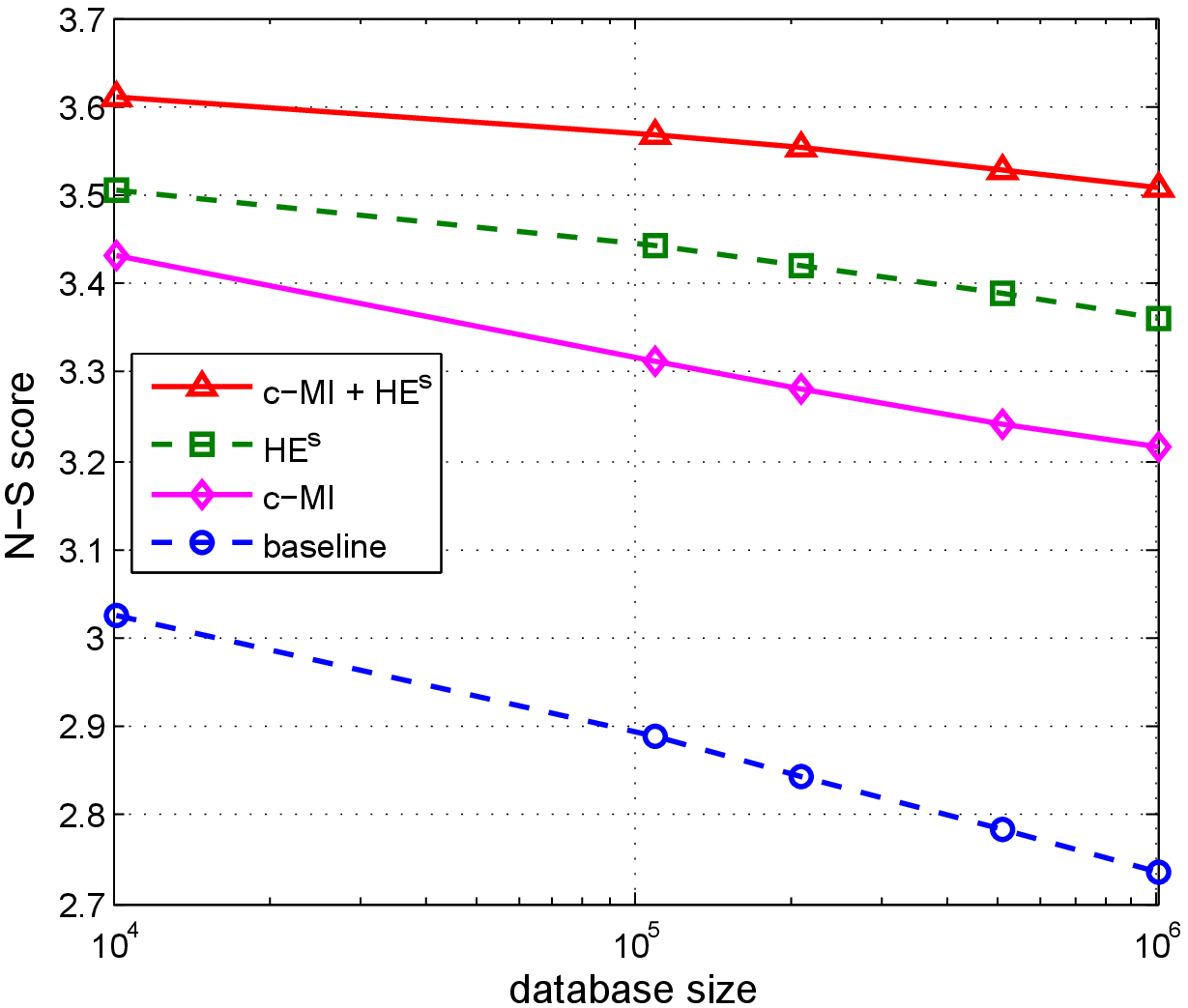}}
\hspace{0in}
 \subfigure[Holidays]{\label{fig:holidays_scale}%
\includegraphics[width=1.6in]{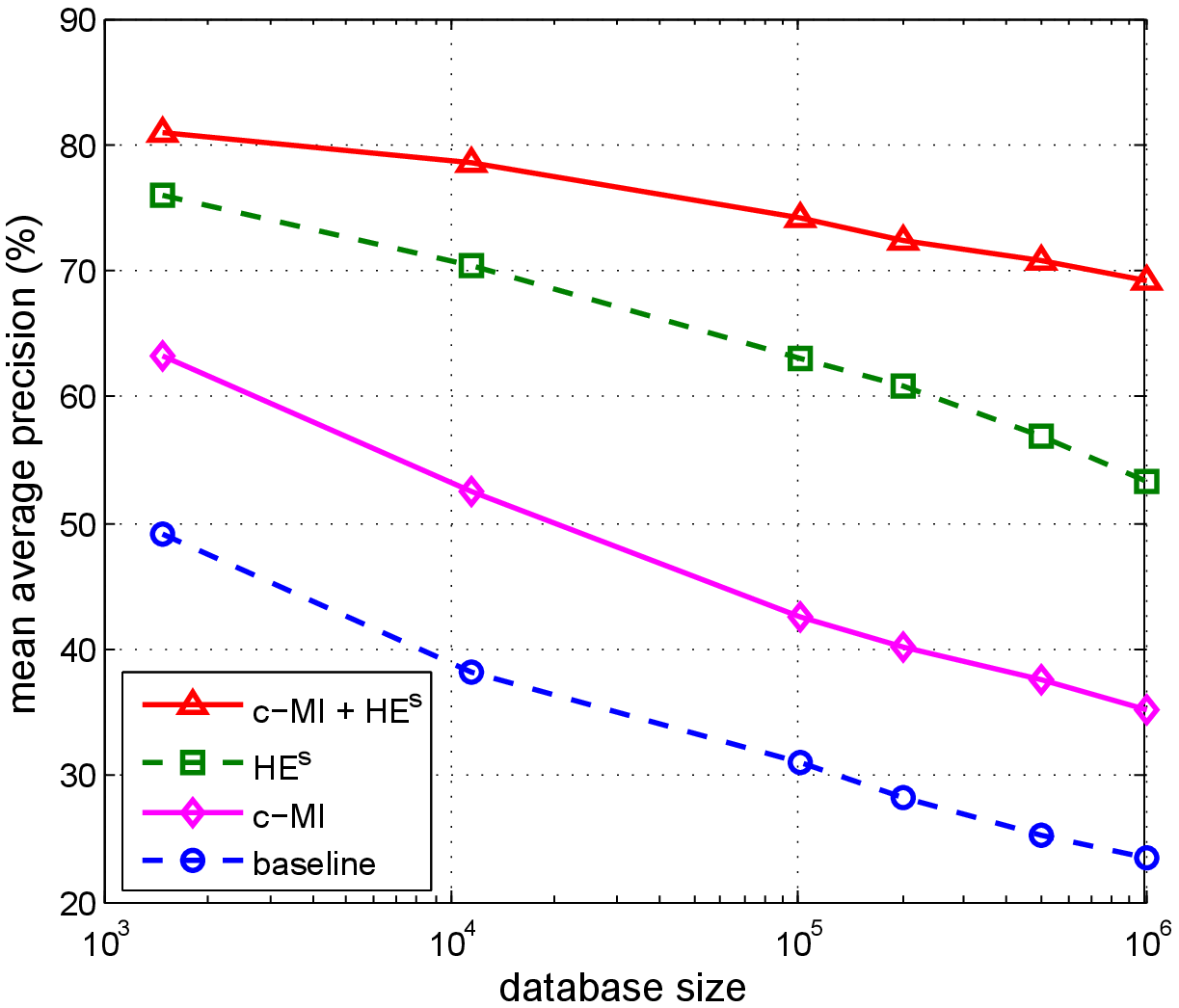}}
\hspace{0in}
 \subfigure[DupImage]{\label{fig:dupimage_scale}%
\includegraphics[width=1.6in]{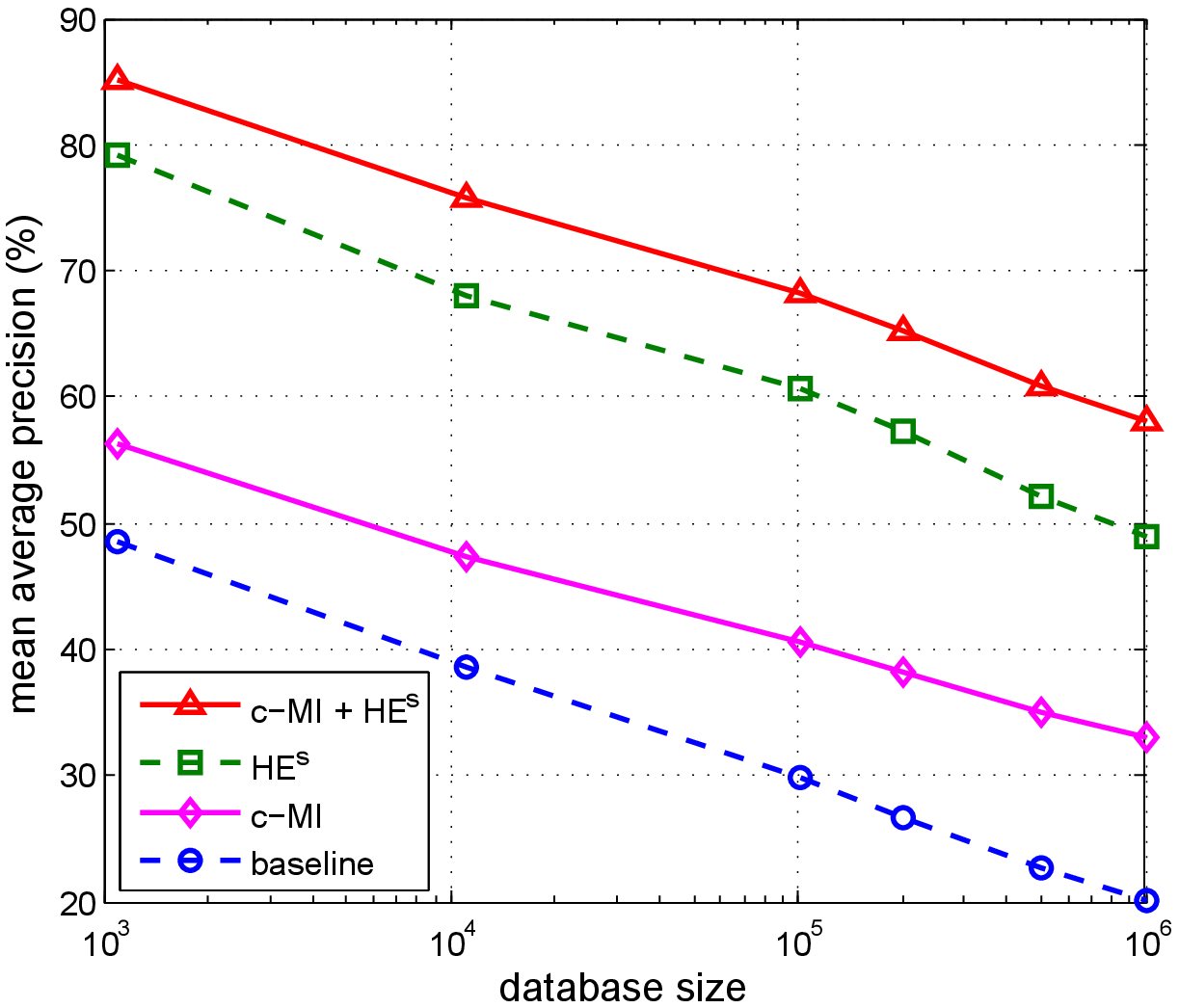}}
\hspace{0in}
 \subfigure[Mobile]{\label{fig:holidays_scale}%
\includegraphics[width=1.6in]{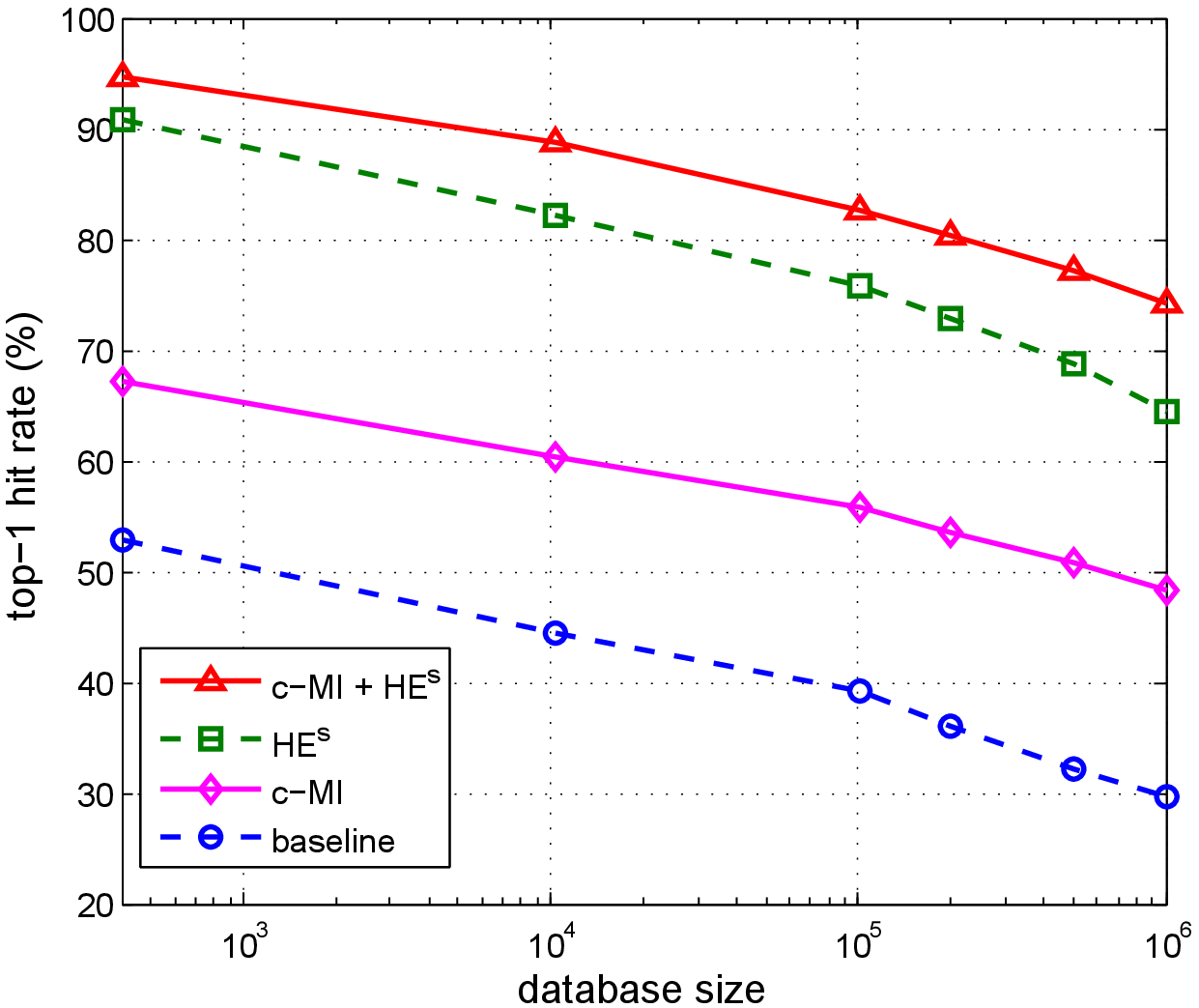}}
\caption{Image retrieval performance against the database size for BoW baseline, c-MI, HE$^s$, and c-MI + HE$^s$ methods.  (a) Ukbench, (b) Holidays, (c) DupImage, and (d) Mobile datasets are merged with various fractions of MIR Flickr 1M dataset.}

\label{fig:mAP-scale}
\end {figure}

\setlength{\tabcolsep}{11.3pt}
\begin{table*}[t]
\centering
\begin{tabular}{|c|ccccccccc|}
\hline
Methods& Ours &  \cite{zhang2012query} & \cite{dengvisual}& \cite{jegou2010accurate} &  \cite{qin2011hello} & \cite{shen2012object} & \cite{jegou2010improving} & \cite{qin2013query}  & \cite{burstiness}\\

\hline
\hline
\emph{Ukbench}, N-S score& \bf 3.85 & 3.77 &84.7& 3.68 &3.67 & 3.56   & 3.55 & - & 3.64\\
\hline
\emph{Holidays}, mAP(\%)& \bf 85.8 & 84.6 &3.75 & - & -   &  -   & 84.8 & 80.1 &  84.8\\
\hline
\end{tabular}
\caption{Performance comparison with state-of-the-art methods with post-processing}
\label{table:state_of_art_post}
\end{table*}
First, we plot the image retrieval accuracy against the database size in Fig. \ref{fig:mAP-scale}. We note that when applied alone, $\mbox{HE}^{s}$ and c-MI each brings about a significant improvement over the baseline. The reason is that both methods works by enhancing the discriminative power of visual words. Moreover, the combination of $\mbox{HE}^{s}$ and c-MI achieves further improvements on all the four datasets. As the database is scaled up, the performance gap between c-MI and the baseline seems to become larger: the feature fusion scheme works better for large databases.

Second, the average query time for the 1M database is presented in Table \ref{table:search_time}. The experiments are performed on a server with 3.46 GHz CPU and 64GB memory. The feature extraction and quantization takes an average of 0.67s and 0.24s on the 1M dataset, respectively. From Table \ref{table:search_time}, the baseline approach is the most time-consuming, e.g. 2.28s for a query in the Ukbench dataset. $\mbox{HE}^{s}$  is more efficient than the baseline due to the filtering effect of the Hamming threshold. On Ukbench, $\mbox{HE}^{s}$ reduces the query time to 1.93s. Furthermore, the c-MI + $\mbox{HE}^{s}$ method proves to be the most time efficient one. On all the four datasets, c-MI cuts the query time to about one half compared to the baseline. The reason lies in that compared with the conventional inverted index, c-MI shortens the list of indexed features per entry. Moreover, since $50\%$ of the color index are traversed, c-MI actually halves the query time. Nevertheless, the query time can be further decreased if fewer entries are visited, i.e., at a cost of lower accuracy.

Third, we discuss the memory cost of c-MI in Table \ref{table:memory}. For each indexed feature, 4 bytes are allocated to store image ID in the baseline. In $\mbox{HE}^{s}$, 8 bytes are needed to store the 64-bit binary SIFT feature. c-MI adds another 22 bits (2.75 bytes) for the binary CN signature. On the 1M dataset, the c-MI + $\mbox{HE}^{s}$ method totally consumes 6.1 GB memory.

\makeatother
The above analysis indicates that c-MI is especially suitable for large scale settings: higher accuracy accompanied with less query time, and acceptable memory cost.

\subsection{Comparison with state-of-the-arts}
We first compare our results with state-of-the-art methods which do not apply any post-processing procedure. As shown in Table \ref{table:state_of_art_no_post}, for Ukbench dataset, we achieve the best N-S score 3.71, which significantly exceeds the result reported in \cite{jegou2010accurate} by +0.10. For Holidays dataset, our result (mAP = 84.0\%) also outperforms the state-of-the-art approaches. By +0.1\% in mAP, our result is slightly higher than \cite{burstiness}. In fact, \cite{burstiness} also employs the inter-image burstiness weighting and weak geometric consistency, which are absent in our retrieval system.

Moreover, in Table \ref{table:state_of_art_post}, we present a comparison with results obtained by various post-processing algorithms, including RANSAC verification \cite{AKM}, kNN re-ranking \cite{shen2012object}, graph fusion \cite{zhang2012query}, and RNN re-ranking \cite{qin2011hello}, etc. We show that, followed by graph fusion \cite{zhang2012query} of modified global HSV feature (HSV*), we have set new records on both Ukbench and Holidays datasets. Notably, we achieve an N-S score of 3.85 on Ukbench, and an mAP of 85.8$\%$ on Holidays, which greatly exceeds the N-S score of 3.77 \cite{zhang2012query} and mAP of $84.8\%$ \cite{burstiness}, respectively. We envision that other post-processing steps can also benefit from our method.

\section{Conclusion}
\label{sectioin: Conclusion}
In this paper, we present a coupled Multi-Index (c-MI) framework for accurate image retrieval. Each keypoint in the image is described by both SIFT and color descriptors. Two distinct features are then \emph{coupled} into a multi-index, each as one dimension. c-MI enables indexing-level feature fusion of SIFT and color descriptors, so the discriminative power of BoW model is greatly enhanced. To overcome the illumination changes and improve recall, a large MA is used for color feature. By further incorporating other complementary methods, we achieve new state-of-the-art performance on Holidays (mAP = 85.8\%) and Ukbench  (N-S score = 3.85) datasets. Moreover, c-MI is efficient in terms of both memory and time (about half compared to the baseline) costs, thus suitable for large scale settings. As another contribution, codes and data are released on our website\footnote{http://www.liangzheng.com.cn}.

In the future, more efforts will be made to explore the intrinsic properties of the coupled multi-index. Moreover, since c-MI can be extended to include other local descriptors, different feature selection strategies and c-MI of higher orders will be investigated.\\

\noindent\textbf{Acknowledgement} First, we would like to thank Dr. Herv\'{e} ~J\'{e}gou for discussion and data sharing in both this paper and \cite{Bayes_merging}. This work was supported by the National High Technology Research and Development Program of China (863 program) under Grant No. 2012AA011004 and the National Science and Technology Support Program under Grant No. 2013BAK02B04. This work also was supported in part to Dr. Qi Tian by ARO grant W911NF-12-1-0057, Faculty Research Awards by NEC Laboratories of America,  and 2012 UTSA START-R  Research Award respectively. This work was supported in part by National Science Foundation of China (NSFC) 61128007.

{\footnotesize
\bibliographystyle{ieee}
\bibliography{egbib}
}

\end{document}